# A Comparative Transfer-Learning Study of CNN Backbones for Partial Face Recognition on the SoF Dataset

Ahmed Kubba
Department of Computer Science
University of Sharjah
Sharjah, UAE
U23103280@sharjah.ac.ae

Ali Alsalama
Department of Computer Science
University of Sharjah
Sharjah, UAE
U23102893@sharjah.ac.ae

Abdelrahman Abdalla
Electrical Engineering and Information Technology
Technical University Dortmund
Dortmund, Germany
abdelrahman.abdalla@tu-dortmund.de

Qassim Nasir
Department of Computer Engineering
University of Sharjah
Sharjah, UAE
nasir@sharjah.ac.ae

Manar Abu Talib
Department of Computer Science
University of Sharjah
Sharjah, UAE
mtalib@sharjah.ac.ae

**Abstract—Face recognition is widely deployed in surveillance, access control, and forensic workflows, yet accuracy degrades sharply once the face is occluded by accessories, foreground objects, or the frame edge. Because most faces in the wild are partial, robust partial face recognition (PFR) remains open. This paper compares three pretrained convolutional backbones, ResNet-50, VGG-16, and FaceNet, fine-tuned for PFR by transfer learning under identical preprocessing, splitting, and optimization protocols on the Specs-on-Faces (SoF) dataset. All three arms use a common 160x160 input and a frozen backbone with a trainable head under a fixed epoch budget and no per-backbone hyperparameter search. The FaceNet configuration, denoted PFN (Partial FaceNet), substantially outperforms the other two, reaching 97.4% test accuracy with macro-averaged 87.04% precision, 84.61% recall, and 84.17% F1 over the 112 identity classes, the highest accuracy and recall reported on SoF. The code is available on this Github Repository.**

## I. Introduction

Face recognition is now embedded in phones, border control gates, and surveillance pipelines. State-of-the-art systems work well on frontal, well-lit, unoccluded faces but degrade sharply once the face is hidden by sunglasses, a mask, a scarf, a hand, or the frame edge [1], [2]. In deployment such conditions are the rule, and the gap between benchmark and field accuracy is largely attributable to occlusion: a classifier can be excellent on the images it is benchmarked on and close to useless on those it is actually shown. Partial face recognition (PFR) targets this gap, identifying a person from whatever portion of the face is visible, with applications in forensics, continuous mobile authentication, attendance, retail analytics, and access control. Interpol reports identifying more than 1,500 individuals of interest since 2016 [3], with field performance constrained almost entirely by partial and degraded input. PFR still trails fingerprint and iris recognition in robustness, and recognizers trained on holistic faces lose accuracy on occluded test inputs [4], [2].

A natural mitigation is to fine-tune a face-pretrained backbone rather than build an architecture from scratch. The practical question is which readily available pretrained CNN transfers best to partial faces, and why. This paper answers it with a matched-protocol comparison of ResNet-50, VGG-16, and FaceNet. The protocol is matched rather than controlled: every arm receives the same data, splits, input tensor, head architecture, and optimizer, but not its own best-case configuration. SoF [5] is used because it couples occlusion (natural eyewear plus synthetic nose and mouth elements) with illumination perturbation. Each model has its top dense layer removed, new classification layers appended, the backbone frozen, and the head trained on the SoF training split. Evaluation uses accuracy, precision, recall, and F1 from the confusion matrix.

Attention- and transformer-based classifiers already outperform both comparison backbones on occluded benchmarks [6], [2], [7], so restricting the comparison to ResNet-50 and VGG-16 may appear to target a superseded generation. This study is not a search for the strongest current PFR model but a two-variable contrast: pretraining corpus varies while architectural family is held approximately constant. The two ImageNet backbones isolate capacity from pretraining domain, since VGG-16 carries 5.4 times the parameters of ResNet-50 (138 M against 25.6 M) and six times those of FaceNet yet shares its pretraining corpus. If capacity was the operative variable, VGG-16 should close part of the gap, and Section IV shows it does not. A Vision Transformer would instead confound pretraining domain with changes of architectural family, inductive bias, and optimization regime, and ViT-scale models [9] are uncompetitive under head-only training at this data scale. These backbones are also the default ImageNet initializations in applied PFR work [8], [2], making a negative result about them directly actionable. Face-pretrained transformers [7] and attention-pooled models [6] are successors to this study rather than omitted baselines.

The contributions of this paper are as follows:

- A matched-protocol evaluation of ResNet-50, VGG-16, and FaceNet as PFR transfer-learning backbones on SoF, with a common 160x160 input below two backbones' pretraining resolution, a fixed epoch budget, and no per-backbone hyperparameter search.
- An analysis of why a face-pretrained backbone transfers better to partial faces than ImageNet-pretrained ones, as evidence about behavior under this protocol.
- A documented, reproducible PFN configuration: stratified splitting across 112 identities, an enlarged batch, a ReduceLROnPlateau scheduler, and the exact checkpoint and corpus.
- A comparison against published SoF baselines showing that PFN attains the best reported accuracy and recall, and the two metrics on which it does not lead, precision and macro-F1.

## II. Literature Review

Work on partial and occluded face recognition splits into three groups: matching local descriptors without prior alignment, localizing and weighting visible regions, and fine-tuning face-pretrained embedding networks. Early local-descriptor work matched a probe to a gallery class by instance-to-class distance. SIFT-based PFR is invariant to misalignment and needs no facial component locations, but hand-crafted descriptors saturate well below deep models and are sensitive to severe occlusion.

Region-based methods localize visible parts before matching. Mahbub et al. [10] clustered facial segments and trained an SVM on statistical features, adding a Deep Regression-based User Image Detector that removes proposal generation, reaching 87.45% true acceptance at 1% false acceptance. He et al. [11] proposed Dynamic Feature Matching, extracting spatial features at the probe's resolution and matching them against a dynamic gallery dictionary for 94.96% accuracy on LFW. Hörmann et al. [6] combined attentional pooling on a truncated ResNet with an aggregation module and adapted losses so attention covers diverse, possibly occluded regions. That line of work, with transformer recognizers [7] and margin-based objectives such as ArcFace [12], is the current frontier, and is outside this comparison because it would confound the pretraining-domain variable isolated here.

The third group, embedding networks fine-tuned for occlusion, is closest to this paper. Mensah et al. [13] found FaceNet embeddings degrade gracefully but predictably under 30% and 40% occlusion with expression variation, motivating fine-tuning over zero-shot use. Hariri [8] cropped to the visible upper face and used a VGG-16 extractor with bag-of-features quantization. Mustapha et al. [14] showed MobileNetV2 supports real-time masked recognition on commodity hardware, and Deng et al. [15] proposed a FaceNet-inspired model 30x smaller that retains discriminability on grayscale occluded inputs.

Two surveys frame the design space. Sharma et al. [17] organize masked face recognition into occlusion-robust feature extraction, occlusion-aware recognition, and occlusion recovery, reporting that face-pretrained backbones consistently outperform ImageNet-pretrained ones on occluded benchmarks. Muhamada et al. [4] concur on a broader benchmark set, and Zhalgas et al. [2] identify attentional re-calibration with region-specific aggregation as promising. Watcharabutsarakham et al. [16] report 92.7% accuracy on mask-occluded faces with YOLOv3 but flag difficulty with eyewear and non-frontal poses. This study quantifies that gap on SoF under one documented protocol, and its size when the ImageNet backbones receive no per-architecture tuning.

Two directions emerge. First, prior knowledge about faces matters on partial inputs: face-pretrained backbones start from feature spaces that already discriminate identity, while ImageNet backbones start from general object features. Second, oscillating validation accuracy is widely reported on small or class-imbalanced PFR datasets and is usually addressed by batch-size adjustment and learning-rate scheduling rather than architectural change.

PFN uses a deep embedding backbone rather than hand-crafted descriptors, starts from FaceNet rather than an ImageNet backbone, and stabilizes the validation behavior reported in [8], [13] through stratified splitting across 112 identities, a ReduceLROnPlateau scheduler, and an enlarged batch. The third is a joint effect: the measures were introduced together in one run and are not separately ablated.

## III. Methodology

### A. Comparative Framework

Three backbones are evaluated under matched conditions: ResNet-50, VGG-16, and FaceNet (Inception-ResNet-v1). For each, pretrained weights are loaded, the top dense layer removed, the remaining layers frozen, and a head appended: 2-D global average pooling, dense layers with ReLU, and a softmax over 112 identity classes. ReLU avoids the vanishing-gradient regime affecting sigmoid and tanh in deeper appended heads [18]. Table I summarizes the backbones. The decisive difference is the pretraining corpus: ResNet-50 [19] and VGG-16 [20] are pretrained on ImageNet, whose 1,000 generic object classes share little with identity discrimination, while FaceNet [21] is pretrained with a triplet loss mapping images of one identity together and different identities apart. The backbone is the Inception-ResNet-v1 of Schroff et al. [21] as distributed in keras-facenet, checkpoint 20180402-114759, trained on VGGFace2 [22] (~3.31 M images, 9,131 identities). The companion 20180408-102900 checkpoint uses CASIA-WebFace and other FaceNet-style models use MS-Celeb-1M [23]. These corpora differ in identity count, per-identity depth, and label noise, so the reported numbers apply to the VGGFace2 checkpoint only.

All three arms receive 160x160 inputs, native to FaceNet, while ResNet-50 and VGG-16 were pretrained at 224x224. Both accept the smaller input, but accepting an input is not the same as being pretrained for it: downsampling partially destroys the fine texture detail the deeper ImageNet blocks detect, so those arms are evaluated outside their pretraining distribution, which penalizes them. The shared input is retained because native-resolution runs would vary resolution together with pretraining corpus, and a common 224x224 would move FaceNet out of distribution instead, exchanging one confound for its mirror image.

TABLE I. THE THREE PRETRAINED BACKBONES AND THEIR ROLES IN THE COMPARATIVE EVALUATION.

| Aspect | ResNet-50 | VGG-16 | FaceNet |
|---|---|---|---|
| Depth (layers) | 50 | 16 | ~22 blocks |
| Approx. parameters | 25.6 M | 138 M | 22.8 M |
| Pretraining Dataset | ImageNet (general) | ImageNet (general) | VGGFace2 |
| Image Input Size | 160x160 | 160x160 | 160x160 |
| Training Objective | Softmax classification | Softmax classification | Triplet loss embedding |
| Per-backbone Tuning | None | None | Batch size, epochs, LR schedule tuned across two runs |

## B. Proposed Architecture

SoF images pass through cleaning and preprocessing, are split with stratification across identities into training, validation, and test sets, and are fed to each backbone in turn. Transfer learning freezes the backbone and appends trainable head layers. Each model is trained, evaluated, and, when its validation curves are unsatisfactory, re-tuned. The loop yields a comparative ranking along accuracy, loss, precision, recall, and F1, from which PFN is selected. One asymmetry should be named: the re-tuning branch was exercised for the FaceNet arm only, with consequences set out in Section IV. Transfer learning is the unifying principle, adapting models trained for generic image classification or face embedding to 112-class PFR on SoF and avoiding retraining of low-level filters [18].

## C. Dataset and Preprocessing

SoF [5] contains 42,592 images of 112 individuals (66 male, 26,112 images, 46 female, 16,480 images), all wearing glasses under varied occlusion and illumination. Built to stress detectors and recognizers with harsh illumination and occlusion, it includes natural occlusion (the eyewear every subject wears), synthetic occlusion (added nose and mouth elements), and three filters designed to evade detection. Images carry subject ID, landmarks, face and eyewear boxes, gender, age, capture year, emotion, and eyewear type, graded easy, medium, or hard.

The 66/46 gender figures are subject-level metadata, not the label space: the models perform 112-way identity classification and gender never enters the loss. The imbalance that matters is the distribution of images per identity, averaging 380 per subject but uneven. Stratified splitting is therefore applied over the 112 identity labels, and the metrics in Section IV are macro-averaged over those classes so a sparse identity counts as much as a dense one.

Preprocessing uses OpenCV. Images are resized to 160x160 to match the FaceNet input and keep the input tensor identical across backbones. Perspective transformation corrects orientation, cropping isolates the face, sharpening enhances detail, and gamma correction normalizes lighting. Images are organized into 112 folders, one per identity, and partitioned 60/20/20 into training, validation, and test sets, stratified by identity. SoF is class-imbalanced, and a random split risks leaving too few instances of some classes to estimate per-class metrics reliably. Augmentation (rotation, shift, shear, zoom, horizontal flip) is applied to the training split only. The split is generated once with a fixed seed and reused across all three backbones, so the arms are compared on identical partitions.

## D. Training Configuration and Hyperparameter Selection

Hyperparameters are identical across backbones except where the FaceNet experiment was iterated to address earlier validation instability. Table II lists each value and its rationale. The FaceNet arm received two runs and three interventions, a doubled batch, a doubled epoch budget, and a reactive scheduler, chosen from its own validation curves. The ResNet-50 and VGG-16 arms received one run each at the shared defaults, with no schedule, no head search, no unfreezing, and no budget extension.

Three interventions would be expected to move these numbers materially in future work: progressive unfreezing of the final residual or convolutional block, which for ImageNet backbones on out-of-domain tasks typically matters more than head design. A learning-rate schedule of the kind the FaceNet arm received, since a fixed 0.001 with Adam on a randomly initialized 112-way head is itself a plausible cause of the non-convergence, and a longer budget, since both arms were still improving when training stopped.

The batch size was raised from 256 to 512 in the final FaceNet run, motivated by the equivalence between a larger batch and a smaller learning rate [24]: doubling the batch halves gradient variance and damps the late-epoch oscillations visible at 256. ReduceLROnPlateau is preferred over fixed decay because the instability is reactive, appearing once the head begins to fit identity-specific cues and then overshoots. Both came from the first run's curves and were introduced together in the second.

TABLE II. EXPERIMENTAL HYPERPARAMETER VALUES.

| Hyperparameter | Value | Rationale |
|---|---|---|
| Input resolution | 160x160 px | Native FaceNet input, identical across all three arms. |
| Batch size (initial) | 256 | Standard for fine-tuning embedding networks at this input size. |
| Batch size (final) | 512 | Halves gradient variance, damped the oscillation seen at 256. FaceNet only. |

| Hyperparameter | Value | Rationale |
|---|---|---|
| Epochs (initial) | 20 (ResNet-50, VGG-16), 25 (FaceNet) | Gauges convergence within a ~5-10 h per-run compute budget. |
| Epochs (final) | 50 | Allows the model to settle after the LR scheduler intervenes. |
| Optimizer | Adam | Adaptive rates suit head-only training. |
| Initial learning rate | 0.001 | Adam default. Frozen base layers preserve learned features. |
| LR scheduler | factor 0.5, patience 3 | Reactive halving counters oscillatory validation curves. |
| Loss function | Sparse cat. cross-entropy | Standard for 112 integer-encoded classes with softmax output. |
| Train/Val/Test split | 60/20/20% (stratified) | Preserves per-class proportions across 112 uneven classes. |
| Random seed / repeats | Fixed seed, 1 run | All figures come from one run on one fixed partition. |

### *E. Evaluation Approach*

Performance is evaluated with four metrics derived from the confusion matrix: accuracy, precision, recall, and F1 [25]. The combination is informative when the cost of a false positive differs from that of a false negative, as in forensic and access-control PFR, and when class proportions are uneven, as on SoF.

Precision, recall, and F1 are macro-averaged over the 112 identity classes: computed per class one-vs-rest, then averaged with uniform weight, so an identity with 150 images counts as much as one with 900. This resists inflation by over-represented classes, and it is why the reported F1 (84.17%) is not the harmonic mean of the reported precision and recall (85.81%): macro-F1 is the unweighted mean of 112 per-class F1 scores. The 2.9-point gap indicates a subset of identities on which precision and recall are jointly poor rather than trading off.

## IV. Results and Discussion

### *A. ResNet-50 Experiment*

The first model uses ResNet-50. With the top dense layer dropped and the rest frozen, a 2-D global average pooling layer and three dense layers are appended, ReLU on the first two and softmax on the last. Training runs 20 epochs with sparse categorical cross-entropy and Adam at a fixed 0.001, no scheduler and no head search, at roughly 12 minutes per epoch for about four hours. In Fig. 1(a) training accuracy improves slowly but steadily while validation accuracy oscillates sharply with only a marginal upward trend, both ending below 0.10. Training loss decreases monotonically but validation loss stays near 5.0 and never converges with it (4.2).

Two readings are available and the data do not separate them. Either ImageNet features provide no usable starting point for 112-class identity discrimination, leaving the head too much of the representational work, or the run is simply undertrained: a fixed 0.001 on a randomly initialized 112-way softmax head is a known recipe for the high-amplitude oscillation in Fig. 1(a), and 20 epochs is a short budget for a head that must learn identity structure from scratch. The decreasing training loss is consistent with both.

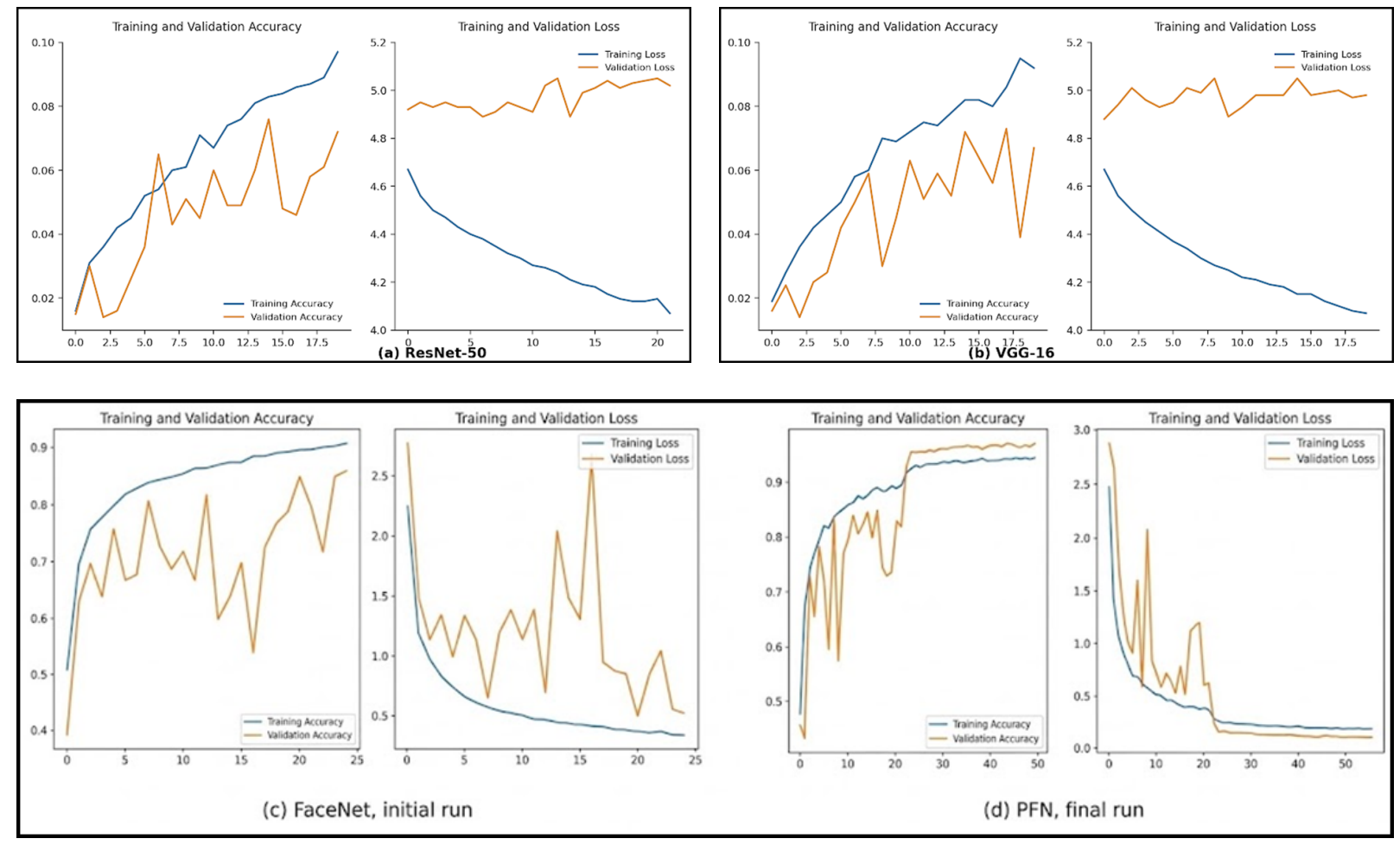

Fig. 1. Training and validation curves for all four runs (x-axis: training epoch, y-axis: metric value). (a) ResNet-50. (b) VGG-16. (c) FaceNet, initial run. (d) PFN, final run.

### *B. VGG-16 Experiment*

VGG-16 is constructed identically, but each epoch takes about 23 minutes, roughly twice the ResNet-50 cost, reflecting its larger parameter count (138 M vs 25.6 M). Fig. 1(b) mirrors Fig. 1(a): training accuracy improves marginally, training loss decreases gradually, and validation accuracy and loss fluctuate without converging. Test accuracy lands around 6%.

The two ImageNet experiments agree, and the form of that agreement is informative: VGG-16 carries 5.4 times the parameters of ResNet-50 and performs no better. Additional capacity brings no improvement here, which argues against an undersized head and for a mismatch between pretraining corpus and task. Under head-only training at 160x160 with shared hyperparameters, ImageNet-pretrained features are not a sufficient starting point for PFR on SoF, and scaling capacity within the ImageNet family does not remedy it.

### C. FaceNet Experiment

FaceNet is loaded with pretrained weights, the embedding layer removed, the rest frozen, and a head of dense layers with LeakyReLU and BatchNormalization appended. The first run trains 25 epochs at batch size 256 with Adam and sparse categorical cross-entropy. In Fig. 1(c) training accuracy rises and loss falls smoothly, confirming the backbone supplies identity-discriminative features. Test accuracy reaches 86%, but validation accuracy and loss oscillate throughout, characteristic of class imbalance with an aggressive learning rate and moderate batch.

The final run extends training to 50 epochs, raises the batch to 512, and adds a ReduceLROnPlateau scheduler halving the learning rate after three stagnating epochs. In Fig. 1(d) training metrics improve steadily, validation oscillations subside once the scheduler intervenes, and both curves converge. Final test accuracy is 97.4% at a loss of about 9.8%, in roughly 10 hours at 12 minutes per epoch, with precision 87.04%, recall 84.61%, and F1 84.17%. The improvement from 86% to 97.4% is the joint effect of those three changes and the doubled epoch budget, applied together in one run.

### D. Validation Fluctuation and Stabilization

The oscillations in Fig. 1 have three causes. SoF's class imbalance and uneven per-identity counts make validation mini-batches sample a shifting distribution. Its adversarial filters and synthetic occlusions introduce large intra-class variation relative to per-class sample count, shifting the decision boundary with whichever perturbed images dominate an epoch. And ImageNet-pretrained backbones supply weak inductive bias for face identity, forcing the head to overfit.

Three remedies were applied jointly, and the stabilization in Fig. 1(d) follows from their combination. The mechanisms below are why each was chosen, not measured contributions. FaceNet replaces the weak ImageNet prior with a face-specific one. Doubling the batch halves gradient variance, reducing its standard deviation by $\sqrt{2}$. ReduceLROnPlateau halves the learning rate on stagnation, shrinking the step precisely where overshooting occurs.

### E. Comparison with Prior Work on SoF

Table III places PFN alongside published SoF baselines. PFN delivers the highest reported accuracy (97.40%) and recall (84.61%), while remaining mid-field on precision (87.04%, exceeded by Zhang et al. and both CDSM variants).

Two qualifications apply. First, the accuracy comparison rests on a single baseline: of the five published results in Table III only AFIF4 [5] reports accuracy. PFN's 97.40% is therefore the highest reported on SoF but is compared against one number rather than the field, and accuracy on a 112-way problem with uneven class sizes is the least discriminating of the four metrics. Second, PFN does not lead on F1: CDSM with SSR attains macro-F1 84.61% against PFN's 84.17%, so on the metric summarizing the precision-recall trade-off the strongest prior baseline is 0.44 points ahead.

PFN nonetheless occupies a distinct point on the precision-recall curve, recovering more true matches than any published SoF system (84.61% recall against 79.66% for CDSM with SSR, 71.12% for Zhang et al., 58.86% for CDSM without SSR, 47.06% for Viola-Jones) at mid-field precision. Whether that trade is favorable is application-dependent: in forensics a missed match ends an investigative line while a false candidate is filtered by a reviewer, favoring recall, whereas in unsupervised access control a false accept is a security failure and CDSM without SSR (95.24% precision, 58.86% recall) is the better choice. PFN extends the available operating points on SoF rather than strictly outperforming them.

Baseline figures are quoted from their sources and were produced under evaluation protocols (splits, preprocessing, in some cases task definitions) that could not be verified. AFIF4 [5] in particular reports gender classification rather than 112-way identity classification, so its 92.10% does not measure the same quantity as PFN's 97.40%. Table III is best read as a summary of what has been reported on SoF, not a controlled benchmark.

TABLE III. PFN AGAINST PUBLISHED SOF BASELINES. "—" MARKS A METRIC NOT REPORTED IN THE CITED PAPER. CDSM IS CASCADED DEFORMABLE SHAPE MODEL, SSR IS SINGLE-SCALE RETINEX. BASELINE FIGURES ARE QUOTED FROM THEIR SOURCES AND WERE OBTAINED UNDER PROTOCOLS DIFFERING FROM THIS STUDY.

| Model | Dataset | Accuracy | Precision | Recall | F1-Score |
|---|---|---|---|---|---|
| PFN (proposed) | SoF | 97.40% | 87.04% | 84.61% | 84.17% |
| AFIF4 [5] (gender classification) | SoF | 92.10% | — | — | — |
| CDSM w/o SSR [5] | SoF | — | 95.24% | 58.86% | 72.76% |
| CDSM with SSR [5] | SoF | — | 90.21% | 79.66% | 84.61% |
| Viola-Jones [26] | SoF | — | 83.47% | 47.06% | 60.19% |
| Zhang et al. [26] | SoF | — | 97.77% | 71.12% | 82.34% |

### F. Limitations and Scope of the Findings

Several design decisions bound what this study establishes. All three backbones run at 160x160, below the 224x224 pretraining resolution of ResNet-50 and VGG-16, so an unknown share of the gap is attributable to resolution rather than pretraining corpus. A resolution-matched replication at 224x224 is needed to separate them. Every result is also from SoF alone: 112 spectacle-wearing identities from one collection protocol, whose occlusions are predominantly eyewear plus synthetic nose and mouth elements, excluding masks, hands, scarves, extreme pose, and surveillance-scale truncation. Evaluation on AR, on LFW with synthetic occlusion, on a masked-face corpus, and ideally on real surveillance footage is required before the finding generalizes.

This paper establishes that on SoF, under head-only transfer at 160x160 with a shared hyperparameter budget, a VGGFace2-

pretrained FaceNet backbone reaches high accuracy where two ImageNet backbones do not, and documents a configuration reproducing that result. It does not establish that ImageNet backbones are unsuitable for PFR in general, that the gap would survive equal tuning or matched resolution, or that the finding transfers to other occlusion types or datasets. The direction agrees with the surveys of [17] and [4]. The main contribution is the quantification and the documented configuration.

## V. Conclusion

This study compared ResNet-50, VGG-16, and FaceNet fine-tuned for partial face recognition on SoF. Under a shared head-only protocol at 160x160, both ImageNet backbones stayed below 10% test accuracy within the allotted budget while FaceNet reached 97.4% accuracy with macro-averaged 87.04% precision, 84.61% recall, and 84.17% F1. PFN attains the highest reported accuracy and recall on SoF, though its macro-F1 sits marginally below the strongest prior baseline, so it extends the available operating points toward higher recall rather than dominating prior work. The evidence supports a conditional claim: when transfer budget is fixed and only a head is trained, the pretraining corpus governs whether a backbone becomes usable far more than depth or parameter count does, VGG-16 carrying 5.4 times the parameters of ResNet-50 and gaining nothing by it. Whether the gap persists once the ImageNet backbones are tuned, unfrozen, and run at native resolution is not established here.

Initial validation instability was resolved through stratified splitting, an enlarged batch, and a reactive LR scheduler, with no architectural change. These were applied jointly and their individual contributions are not separated. Future work should address the limitations above: an equal-effort baseline giving ResNet-50 and VGG-16 per-architecture schedules, progressive unfreezing, and a head search at native 224x224, multi-seed repetition reporting mean and standard deviation, with a factorial ablation over scheduler, batch size, and epoch budget. Evaluation on AR, LFW with synthetic occlusion, and a masked-face corpus, attention- and transformer-based backbones, including face-pretrained transformers [7] and the attention-pooled models of Hörmann et al. [6], and metric-learning heads such as ArcFace [12] to close the precision gap without sacrificing recall.